\documentclass{article}
\usepackage{spconf,amsmath,graphicx}
\usepackage{graphicx} 
\usepackage{xcolor}
\usepackage{subfigure} 
\usepackage{array}
\usepackage[numbers]{natbib}
\setcitestyle{square}

\usepackage{algorithm}
\usepackage{algorithmic}

\usepackage{hyperref}

\usepackage{xcolor}
\usepackage{enumerate}
\usepackage{diagbox}
\usepackage{enumitem}
\usepackage{multirow}
\usepackage{comment}
\usepackage{lipsum}
\usepackage{booktabs}
\usepackage[utf8]{inputenc}
\usepackage{color}
\usepackage{algorithm}
\usepackage{amssymb}
\usepackage{amsmath}
\usepackage{bm}

\usepackage{algorithm}

\newcolumntype{A}{>{\centering\arraybackslash}m{6ex}}

    \title{RNN-T for Latency Controlled ASR WITH IMPROVED BEAM SEARCH}
\name{
\begin{tabular}{c}
Mahaveer Jain, Kjell Schubert, Jay Mahadeokar, Ching-Feng Yeh,
\\
Kaustubh Kalgaonkar, Anuroop Sriram, Christian Fuegen, Michael L. Seltzer
\end{tabular}
}
\address{Facebook AI, USA}
\begin{document}
%

\maketitle
\begin{abstract}
Neural transducer-based systems such as RNN Transducers (RNN-T) for automatic speech recognition (ASR) blend the individual components of a traditional hybrid ASR systems (acoustic model, language model, punctuation model, inverse text normalization) into one single model. This greatly simplifies training and inference and hence makes RNN-T a desirable choice for ASR systems. In this work, we investigate use of RNN-T in applications that require a tune-able latency budget during inference time. We also improved the decoding speed of the originally proposed RNN-T beam search algorithm. We evaluated our proposed system on English videos ASR dataset and show that neural RNN-T models can achieve comparable WER and better computational efficiency compared to a well tuned hybrid ASR baseline.





\end{abstract}
\begin{keywords}
RNNT, LC BLSTM, E2E, ASR.
\end{keywords}
\section{Introduction}
\label{sec:intro}

Automatic speech recognition (ASR) with Deep Neural Networks (DNN) operates
in a hybrid framework using several models. These models include: 
DNN acoustic models (AM) that estimate the posterior
probabilities of Hidden Markov Model (HMM) states, language models (LM) that estimate probabilities of word sequences, punctuation models and inverse text normalization (ITN) models dealing with number \& date formatting. 
These models are optimized independently~\cite{hinton2012deep} and then combined together using Weighted Finite State Transducer (WFST) for efficient decoding.

End-2-End(E2E) speech recognition techniques such as connectionist temporal classification (CTC) ~\cite{graves2006connectionist}, listen, attend and spell(LAS) ~\cite{las} and RNN-T~\cite{rnnt-graves, rnnt_arch, streaming_rnnt} have become successful because of 
advances in neural networks to model context and history in audio and text  sequences~\cite{sak2014long}. E2E speech recognition combines all components of hybrid ASR model such as AM, LM, punctuation model and ITN into one component and predicts words directly from input acoustics. 

E2E simplifies the training process for a new ASR system, but in order to run them in a server-side application they need to meet the following constraints: 1) streamable with constrained latency 2) match or improve the computational efficiency and WER of the baseline hybrid system. 

LC BLSTM~\citep{lcblstm2, lcblstm_xue} are widely used to build ASR systems with constrained latency. They achieve this by using bi-directional context within short audio chunks without consuming the whole utterance. In this work, we use LC BLSTM for the Audio Encoder of RNN-T to achieve streamable ASR.  
Popular hybrid ASR decoding techniques such as static decoder \cite{povey2011kaldi}  and dynamic decoder \cite{dynamic_decoder} use hyper parameters (beams) to prune hypotheses to  improve  computational efficiency.  Inspired by these works we modify the RNN-T beam search to make it computationally more efficient.
We evaluate our model under various settings for latency control using `throughput`, defined as the number of audio seconds processed per wall clock second on a fixed server CPU architecture, and rtf@40, defined as real time factor at 40 concurrent audio streams.




The rest of the paper is organized as follows. In Section~\ref{sec:RNN-T}, we review the RNN-T model and the LC-BLSTM layer. We present our proposed changes in RNN-T beam search to improve computational efficiency in Section ~\ref{sec:beamsearch}. We discuss our experimental setup and summarize our findings in Section~\ref{sec:exp}. Finally, we conclude with a discussion of future work in Section~\ref{sec:conclusion}.

\section{RNN Tranducer}
\label{sec:RNN-T}

\begin{figure}[t]
  \centering
  \includegraphics[width=\linewidth]{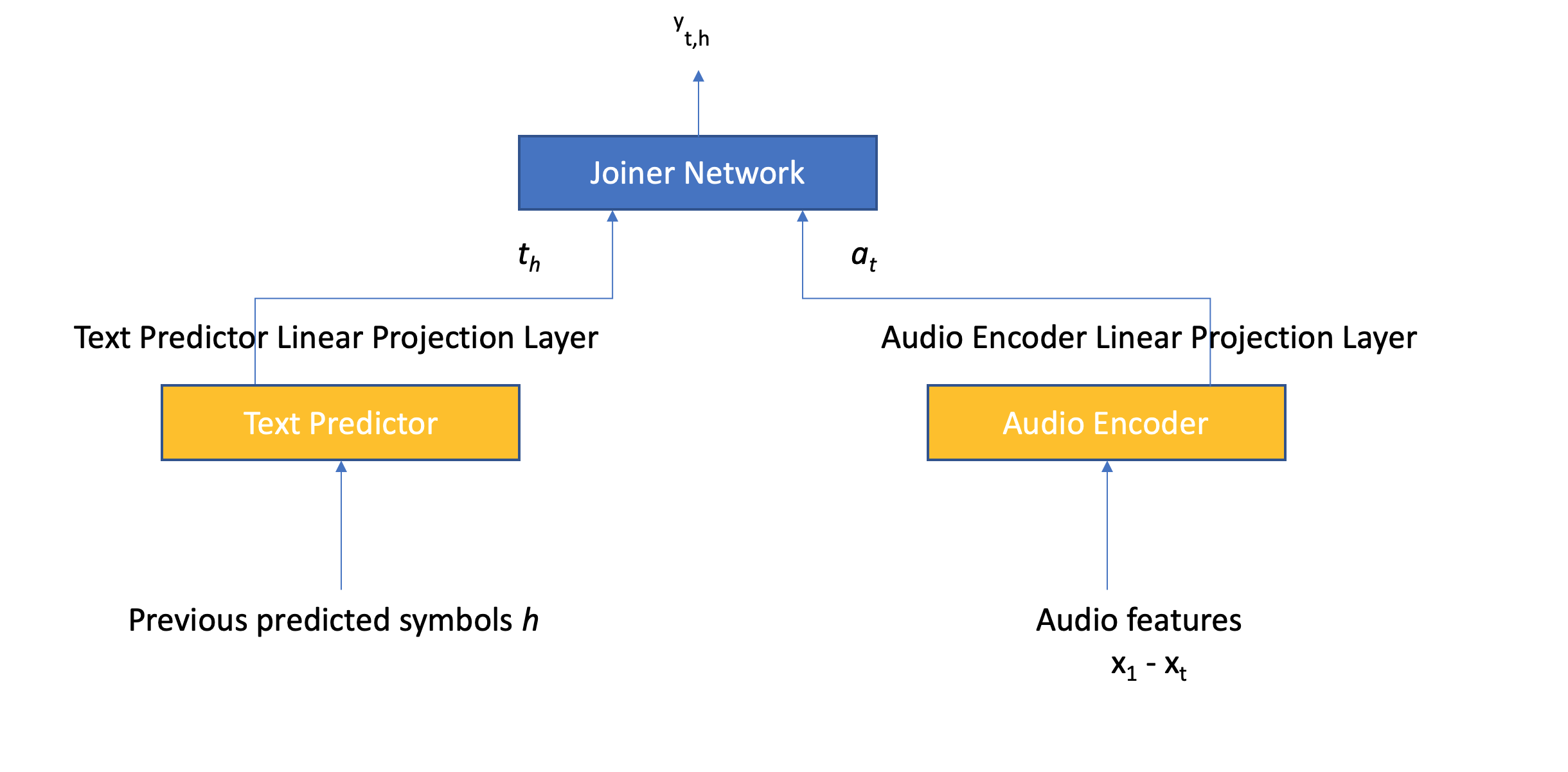}
 \caption{RNN Transducer for ASR}
  \label{fig:rnnt_fig} 
\end{figure}

The framework of RNN-T ASR system is illustrated in Fig.~\ref{fig:rnnt_fig}. RNN-T for ASR has three main components: Audio Encoder, Text Predictor and Joiner. The Audio Encoder encodes audio frames up to a time $t$ as audio embedding $a_{t}$. The Text Predictor encodes the text history up to index $h$ in reference or hypothesis as text embedding $t_{h}$. These embeddings are then fed to the Joiner which combines them to produce a probability distribution over the output units at $y_{t,h}$. By incorporating both audio and text for producing probabilities over output symbols RNN-T can overcome the conditional independence assumptions of CTC models~\cite{graves2006connectionist}. In RNN-T, the output units include a special $blank$ symbol to decide whether to move to next time frame $t+1$ or to emit more output units from same time frame $t$ for the next Joiner call. After every Joiner call we either move in time(t) axis to process next audio frame $t+1$ or we update the hypothesis ($h$) and emit more symbols from the same time frame $t$. The former is done when the Joiner emits a $blank$ symbol, whereas the latter is done for non $blank$ emission.  

\subsection{Latency Controlled BLSTM for RNN-T}
\label{subsec:lcblstm_model}
\begin{figure}[t]
  \centering
  \includegraphics[width=\linewidth]{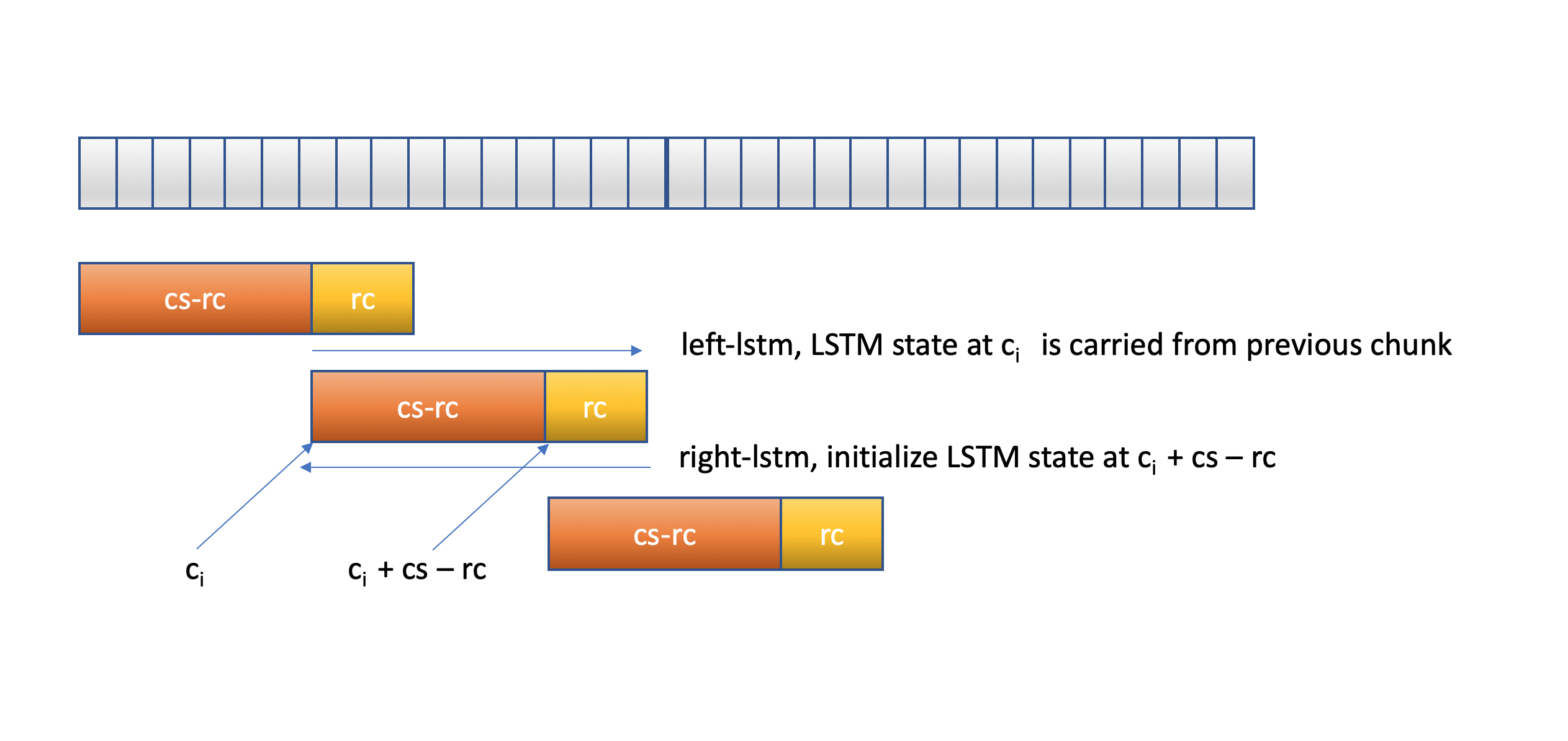}
 \caption{LC BLSTM for RNN-T}
  \label{fig:lcblstm_fig} 
\end{figure}

Unidirectional Audio Encoder models such as LSTMs base their predictions only on the audio history to the left and thus tend to yield worse word error rates than bi-directional encoders that have full left and right context. For live streaming application, we are constrained to use unidirectional encoders because transcript should be made available with minimum possible delay as audio is fed in.
However, some applications permit a certain maximum latency between consuming parts of the input audio and producing the transcript for it. In such cases it greatly helps to use some amount of right context in the Audio Encoder to improve WER. Traditional BLSTM can not produce a transcript until the whole audio stream is processed.  LC BLSTM~\cite{lcblstm_xue, lcblstm2} allows streamable application that has constrained latency. LC BLSTM (fig \ref{fig:lcblstm_fig}) has two LSTMs, left-lstm that runs from left to right in time axis whereas right-lstm runs from right to left in time axis. 

In order to run LC BLSTM for RNN-T, the audio sequence is first divided into overlapping chunks of size $cs$. The amount of overlap between chunks is equal to the minimum amount of right context ($rc$) available to frames in the chunk. As shown in fig \ref{fig:lcblstm_fig}, the amount of right context available is maximum ($cs$) at the first frame of the chunk and it reduces linearly to $rc$ at the end of chunk at $cs-rc$.
This allows every frame in the chunk to have some amount of right context to generate a high-quality audio embedding without delaying the generation of embedding until the whole audio stream is ingested.



\section{Improving Beam search for RNN-T}
\label{sec:beamsearch}

\title{Euclid's Algorithm: An example of how to write algorithms in \LaTeX{}}

\author{write\LaTeX{}}

\date{\today}



        


Inspired by speed improvements in decoders such as Kaldi \cite{povey2011kaldi} using pruning, we modify the RNN-T beam search described in \cite{rnnt-graves} to prune unlikely paths early during decoding to improve `throughput` and real time factor. Our modified algorithm is presented in Algorithm \ref{rnnt_beamsearch_algo}. We use same symbols as used in Algorithm 1 from \cite{rnnt-graves}. 

\begin{algorithm}[t] 
\caption{Improved RNNT Beam Search} 
\label{rnnt_beamsearch_algo} 
Proposed modifications in Algorithm 1 from \cite{rnnt-graves} are in \textcolor{red}{red}.

\begin{algorithmic} 
\STATE \textbf{Initialize:} \hspace{0.05in} $B$ = $\{\varnothing\}$; \hspace{0.05in} $Pr(\varnothing)$ = 1
\FOR {$t$ = 1 to $T$}
    \STATE $A = B$
    \STATE $B = \{\}$
    \FOR {$y$ in $A$}
    \STATE $ Pr(y) += \sum_{\hat{y}\in pre f(y)\cap A} Pr(\hat{y}) Pr(y \mid \hat{y},t) $
    \ENDFOR
    \WHILE{$B$ contains less than W elements more
probable than the most probable in $A$}
        \STATE $y^\ast = $ most probable in $A$ 
       \STATE
       \textcolor{red} { $a\_best\_prob =$ max probability in $A$
        \STATE  {$b\_best\_prob =$ max probability in $B$ }}
        \IF{\textcolor{red}{$log(b\_best\_prob) \ge state\_beam + log(a\_best\_prob)$}}
         \STATE \textcolor{red}{\textbf{break} \COMMENT{meet beam search and break \textbf{while} loop}}
        \ENDIF
        \STATE Remove $y^\ast$ from $A$
        \STATE $Pr(y^\ast) = Pr(y^\ast) Pr(\varnothing \mid y,t)$
        \STATE Add $y^\ast$ to $B$
            \STATE $best\_prob = \max_{k \in non\_blank} Pr(k \mid y^\ast,t)$
            \FOR{$k \in Y$}
            
           \IF{\textcolor{red}{$ log(Pr(k \mid y^\ast,t)) \ge log(best\_prob) - expand\_beam$}}
                \STATE \textcolor{red}{$Pr(y^\ast + K) = Pr(y^\ast) Pr(k \mid y^\ast,t)$}
                \STATE \textcolor{red}{Add $y^\ast + k$ to $A$}
            \ENDIF
             \ENDFOR
        
    \ENDWHILE
    \STATE Remove all but the $W$ most probable from $B$
\ENDFOR
\RETURN  $y$ with highest log $Pr(y)/|y| $ in $B$ 
\end{algorithmic}
\end{algorithm}

In order to explain RNN-T beam search let us assume that hypothesis $h$ at time $t$ has audio embedding $a_{t}$ and text embedding $t_{h}$. These embeddings are fed to the Joiner which combines them to produce probabilities over output units at $y_{t,h}$. As explained in section ~\ref{sec:RNN-T} output units include a special symbol $blank$ ($\varnothing$) to decide whether to move to the next time frame $t+1$ or to emit more output units from the same time frame $t$. In order to ensure that at-least  top $W$(beam size) hypotheses that are being moved to $t+1$ have higher probability than the ones that can still be generated from $t$, a beam search is performed using two sets of hypothesises, $A$ and $B$.

Set $A$ contains hypothesises that are still being considered for time $t$ whereas hypothesis set $B$ contains hypothesises that have already emitted a $blank$ symbol at time $t$, and are now in time frame $t+1$.  As soon as $B$ has $W$ hypotheses more probable than the most probable hypothesis in $A$ the beam search criterion is met at time $t$ and we can start processing frame $t+1$. During beam search we pick the best hypothesis in $A$ and expand it either with $blank$ or non $blank$ symbols. The expansion with $blank$ moves a hypothesis to $B$ whereas expansions with non $blank$ symbols are put back in $A$, which results in a expanded set $A$ at $t$. We introduce $expand\_beam$ to limit number of expanded hypothesises that are added in $A$. For a Joiner call at ($t$,$h$) that produces $y_{t,h}$ we first compute the best prob, $best\_prob$ among non $blank$ output units($y_{t,h}$) and only consider output units($k$) that have higher $ log(Pr(k \mid y^\ast,t)$ than $log(best\_prob) - expand\_beam$ to be added to $A$.

We also introduce a $state\_beam$ and use it as an additional hyper parameter of the beam search. If the best hypothesis in $A$ is worse by more than $state\_beam$ from the best hypothesis in $B$ in log space, we assume that future expansions of hypothesises available in $A$ are too unlikely to compete with already existing hypos in $B$. We always use the natural logarithm of numerical value while discussing $state\_beam$ and $master\_beam$ in rest of the paper.

In experimentation section we show that we can improve `throughput` from 53 to 65 and decrease rtf@40 from .75 to .60 by using $state\_beam$ and $master\_beam$ with negligible WER impact.



\section{Experiments}
\label{sec:exp}

\subsection{Dataset}
\label{subsec:dataset}
The dataset used for our experiments was sampled from English videos shared publicly on Facebook. The data does not contain any
user-identifiable information and is completely anonymized . The training set consists
of around 1M videos with ~13.7K hours in total. We use two
test sets; vid-clean and vid-noisy. Vid-clean has 1.4K videos (about 20.9 hours) whereas vid-noisy that is more acoustically challenging has 1.3K videos (about 20.1 hours). More information about our data sets can be found in \cite{base_hybrid_training}.

 
\begin{table}[thbp!]
    \centering
 \vspace{-0.8em}
  \caption{\label{tab:rep-train-utterances} {\it Amount of training/test data in hours} }
    \begin{tabular}{ | c |  c | c | c  | c | c|  }
    \hline
       \ dataset   & English(hours) \\ 
    \hline
    train & 13.7k \\
    \hline
    vid-clean & 20.9  \\
    \hline
    vid-noisy & 20.1  \\
    \hline
    \end{tabular}
  \label{tab:rep-train-utterances}
  \vspace{-0.8em}
\end{table}

\subsection{Model}
\label{subsec:model}
The architecture of the RNN-T model (Figure \ref{fig:rnnt_fig}) used for the experiments in this paper is as follows. The Audio Encoder has two components: a 5-layer LC BLSTM with 704 dimensions and Audio Encoder Linear Projection Layer (AELPL) of dimension $704$ by $704$. We use subsampling of 2 across the time dimension after the first LC BLSTM layer to improve training and inference speed. The LC BLSTM uses a right context ($rc$) of 20 frames (200ms) and chunk size ($cs$) of 240 frames (2400ms) during training. The Text Predictor also has two components; a 2-layer LSTM of 704 dimensions and Text Predictor Linear Projection Layer (TPLPL) of dimension $704$ by $704$. The Joiner uses concatenation of three layers: summation layer,  ReLU \cite{relu} layer and, a softmax layer and produces probabilities over output units ($y_{t,h}$). We used a token set consisting of 200 sentence pieces, learnt using the sentence piece library \cite{sentence_piece}. The entire model consists of 62M parameters. 

The input to the network consists of globally normalized 80-dimensonal log Mel-filterbank, extracted with 25ms FFT windows and 10ms frame shifts. We use the Adam optimizer\cite{kingma2014adam}, learning rate of 0.0004, with dropout probability of 0.3 and policy LB of SpecAugmentation\cite{spec_aug} during training. Dropout is applied in all layers of LC BLSTM of Audio Encoder and LSTM layers of Text Predictor. 
The RNN-T training was ran for 25 epochs. 

The latency budget can be chosen by setting the Decoding Threshold ($DT$) at inference time. Decoding Threshold($DT$) is defined as the chunk size (in milliseconds) used during inference for LC BLSTM Audio Encoder. If not explicitly specified, we use $DT$ of 800ms for our experiments. We use same amount of right context ($rc$) during training and inference. 

A beam size of 5, expand\_beam of 2.3 and state\_beam of 4.6 were used during inference. INT8 quantization from Pytorch 
was used during inference to speed up decoding. RNN-T is fully neural and does not use an external LM.

The baseline hybrid ASR \cite{base_hybrid_training} system consisted of a 5-layered LC BLSTM model with 800 hidden units and an external WFST language model trained using transcripts from the same training data. The hybrid ASR system was trained using the model and policy described in \cite{base_hybrid_training} to minimize the cross-entropy (20 epochs) loss first, followed by the LF-MMI criterion (8 epochs) \cite{lffmi}. The hybrid ASR system also used INT8 quantization 
during inference time to speed up decoding.
\subsection{Impact of expand and state beam on WER and Throughput}
\label{sec:expand_beam_res}
As discussed in Section \ref{sec:beamsearch}, the introduction of $expand\_beam$ and  $master\_beam$ in beam search of RNN-T allows us to limit the number of hypotheses in the sets $A$ and $B$ at inference time. This boosts the `throughput` by around 22 percent and decrease rtf@40 by 20 percent with negligible impact on WER. We achieve a `throughput` of 65 and rtf@40 of .60 with $expand\_beam$ = 2.3 and $state\_beam$ = 4.6, compared to a `throughput` of 53 and rtf@40 of .75 without the improved beam search. Table \ref{tab:beam_impact} shows the WERs for the vid-noisy test set with different values of these parameters.

\begin{table}[thbp!]
\centering
 \vspace{-0.6em}
  \caption{\label{tab:beam_impact} {\it Impact of beams on WER / Throughput / rtf@40 for vid-noisy} }
    \begin{tabular}{ | A | A | c | c  | c |  }
    \hline
       Expand Beam  & State Beam  & WER  & Throughput & rtf@40 \\ 
    \hline
    1.5 & 2.3 & 21.8 & 68  & .58 \\
    \hline
    2.3 & 4.6 & 21.0 & 65  & .60 \\
    \hline
     inf &  inf & 21.0 & 53  & .75 \\
    \hline
    \end{tabular}
  \label{tab:num-entity-train}
  \vspace{-1em}
\end{table}

\subsection{Comparing Hybrid ASR Model with RNN-T}
\label{sec:exp-hybrid_comp}

At 65MB, the RNN-T model is more than $10\times$ smaller than hybrid ASR baseline model\cite{base_hybrid_training}, while obtaining a comparable WER. In addition to being larger, the Hybrid ASR model also requires its various components, such as the acoustic model, language model, punctuation model and inverse text normalization to be trained individually. Each of these components has their own data and training pipelines that need to be maintained separately. The RNN-T model combines these components into a single model that can be trained end-to-end, which simplifies the training and deployment process.

As seen in the Table \ref{tab:comp_rnnt_hybrid_wer}, RNN-T achieves a similar WER and better `throughput` and rtf@40 for both test sets compared to hybrid system. $DT$ of 800ms was used both for hybrid and RNN-T ASR system.


\begin{table}[thbp!]
\centering
 \vspace{-0.5em}
  \caption{\label{tab:comp_rnnt_hybrid_wer} {\it Comparison of Hybrid model with RNN-T} }
    \begin{tabular}{ | c | c |  c | c | c| }
    \hline
       \ Test Set & System  & WER & Throughput & rtf@40
       \\ 
    \hline
    vid-clean & hybrid & 14.0 & 55 & .70 \\
    \hline
    vid-clean & RNN-T & 14.0 & 63 & .60  \\ 
    \hline

    vid-noisy & hybrid & 20.7 & 55 & .71 \\
    \hline
    vid-noisy & RNN-T & 21.0 & 65 & .60   \\ 
    \hline

    \end{tabular}
  \label{tab:comp_rnnt_hybrid_wer}
  \vspace{-1em}
\end{table}


\subsection{Impact of Decoding Threshold on WER and Throughput}
\label{sec:exp-lcblstm}
The latency budget of ASR systems varies depending on the application that they are being used for. Chunk size of LC BLSTM layer used at inference time can be different from training chunk size ($cs$). We defined Decoding Threshold ($DT$) as the chunk size used during inference time in Section  \ref{subsec:model}. $DT$ parameter gives us a way to achieve flexible latency budget.
By adjusting the $DT$ parameter at inference time, we can trade-off the latency (and throughput) of the system with WER. For larger values of $DT$, the latency between the input audio and the produced transcript is larger, but the system can achieve a better WER, because the average amount of right context available per frame increases. For smaller values of $DT$, the models needs to perform more computations per time step as the right context $rc$ of each audio chunk has to be re-processed for the subsequent chunk, which reduces throughput.

As seen in table \ref{tab:lc_blstm_blstm_Wer_noisy}, the `throughput` decreases from 74 to 48 and rtf@40 increases from .53 to .81 when the $DT$ is decreased from 2000 to 300 for the vid-noisy data set. For our model, the WER increases by 8.9\% (relative) for vid-noisy and 13.3\% (relative) for vid-clean when the $DT$ is decreased from 2000 to 300 as observed in Table \ref{tab:lc_blstm_blstm_Wer_noisy}. We only show  `throughput` and rtf@40 for vid-noisy test in Table \ref{tab:lc_blstm_blstm_Wer_noisy}, vid-clean follows a similar pattern. 

RNN-T models can also be made streamable by using only uni-directional LSTMs in the Audio Encoder. However, our best unidirectional RNN-T only achieves a WER of 16.4\% for vid-clean and 23.6\% for vid-noisy. 
Tuning $DT$ at inference time allows us to get better WERs than with unidirectional models while keeping ASR streamable.

\begin{table}[thbp!]
\centering
 \vspace{-0.5em}
  \caption{\label{tab:lc_blstm_blstm_Wer_noisy} {\it WER and computational efficiency for different $DT$} }
    \begin{tabular}{  |c | c| c | c | c | }
    \hline
        $DT$ & vid-noisy & vid-clean & Throughput & rtf@40 \\ 
      \hline
    2000 & 20.3 &  13.5 & 75 & .53 \\
    \hline
    1500 &  20.4 & 13.5 & 70 & .56 \\
    \hline
    800 &  21.0 & 14.0 & 65  & .60\\
    \hline
    400 & 21.7 & 14.8 & 53 & .74  \\
    \hline
    300 & 22.1 & 15.3 & 48 & .81  \\
    \hline
    \end{tabular}
  \label{tab:lc_blstm_blstm_Wer_noisy}
  \vspace{-1em}
\end{table}

\section{Conclusion}
In this work we show that RNN-T systems is suitable for streaming ASR with latency constraints. Our experiments demonstrate that RNN-T can achieve a good trade-off between latency and WER with LC-BLSTM. 
Our work improves on existing work in two ways: first, the changes we propose to the beam search procedure improve rtf@40 by relative 20\% without impacting WER; second, we show that we can achieve a better WER with an RNN-T equipped with LC-BLSTM layers than one with only unidirectional LSTMs, while still keeping it streamable. The use of LC-BLSTMs also allows the latency of the models to be controlled at inference time. Future directions include using contextual information for RNN-T.
\label{sec:conclusion}



\section{Acknowledgement}
Authors would like to thank Awni Hannun and Yun Wang for the discussions and suggestions about this work.


\bibliographystyle{IEEEbib}
\bibliography{refs}

\end{document}